\pdfoutput=1

\documentclass[11pt]{article}

\usepackage[]{EMNLP2023}

\usepackage{times}
\usepackage{latexsym}

\usepackage[T1]{fontenc}

\usepackage[utf8]{inputenc}

\usepackage{microtype}

\usepackage{inconsolata}

\usepackage{graphicx}
\usepackage{subfigure}
\usepackage{svg}
\usepackage{CJKutf8}
\usepackage{setspace}
\usepackage{multirow}
\usepackage{subfloat}
\usepackage{graphicx}
\usepackage{url}
\usepackage{tikz}
\usepackage{pgfplots}
\usepackage{color}
\usepackage{mathtools}
\usepackage{makecell}
\usepackage{colortbl,booktabs}
\usepackage{paralist}
\usepackage{amsmath,amssymb,bm}
\usepackage{threeparttable}
\usepackage{arydshln}
\usepackage{soul}
\usepackage{changes}
\usepackage{bbding}

\title{SkillNet-X: A Multilingual Multitask Model with Sparsely Activated Skills}

\author{
Zhangyin Feng \textsuperscript{\rm 1},
Yong Dai, 
Fan Zhang, 
Duyu Tang, \\
\textbf{Xiaocheng Feng} \textsuperscript{\rm 1,2}, 
\textbf{Shuangzhi Wu}, 
\textbf{Bing Qin} \textsuperscript{\rm 1,2}, 
\textbf{Yunbo Cao}, 
\textbf{Shuming Shi} \\
$^1$Harbin Institute of Technology, China \\
$^2$Peng Cheng Laboratory, China \\
\{zyfeng, xcfeng, qinb\}@ir.hit.edu.cn 
}

\begin{document}
\maketitle
\begin{abstract}
Traditional multitask learning methods basically can only exploit common knowledge in task- or language-wise, which lose either cross-language  or cross-task knowledge.
This paper proposes a general multilingual multitask model, named SkillNet-X, which enables a single model to tackle many different tasks from different languages.
To this end, we define several language-specific skills and task-specific skills, each of which corresponds to a skill module. 
SkillNet-X sparsely activates parts of the skill modules which are relevant either to the target task or the target language. Acting as knowledge transit hubs, skill modules are capable of absorbing task-related knowledge and language-related knowledge consecutively.
Based on Transformer, we modify the multi-head attention layer and the feed forward network layer to accommodate skill modules.
We evaluate SkillNet-X on eleven natural language understanding datasets in four languages.
Results show that SkillNet-X performs better than task-specific baselines and two multitask learning baselines (i.e., dense joint model and Mixture-of-Experts model).
Furthermore, skill pre-training further improves the performance of SkillNet-X on almost all datasets. 
To investigate the generalization of our model, we conduct experiments on two new tasks and find that SkillNet-X significantly outperforms baselines.
\end{abstract}

\section{Introduction}

Multitask learning (MTL) has been used successfully in a wide range of natural language processing applications by exploiting common knowledge among tasks \cite{ruder2017overview, liu-etal-2019-multi, rahimi-etal-2019-massively}.
The existing multitask learning is mainly implemented in two ways: (1) task-wise multitask learning: different tasks learning in the same language, such as MT-DNN \cite{liu-etal-2019-multi}, which learns representations across multiple English natural language
understanding tasks, and (2) language-wise multitask learning: the same task learning in different languages, such as MW-MNMT \cite{firat-etal-2016-multi}, which enables a single neural translation model to
translate between multiple languages.

\begin{table}[t]
\centering
\begin{tabular}{ccc}
\toprule
\bf Approach & Cross-task & Cross-language \\
\midrule
MT-DNN  & \Checkmark & \XSolidBrush \\
MW-MNMT & \XSolidBrush & \Checkmark \\
SkillNet-X&\Checkmark &  \Checkmark \\
\bottomrule
\end{tabular}
\caption{Comparision of multitask learning approaches. MT-DNN (task-wise) and MW-MNMT (language-wise) are only cross-task and cross-language, respectively, whereas SkillNet-X is both cross-task and cross-language.}
\label{tab:motivation}
\end{table}

As shown in Table~\ref{tab:motivation}, task-wise MTL exploits knowledge from multiple tasks in a same language (i.e., cross-task knowledge) and language-wise MTL benefits from knowledge from a same task in different languages (i.e., cross-language knowledge). 
However, task-wise and language-wise MTL are unable to utilize simultaneously cross-task and cross-language knowledge from different tasks and languages. 
In practice, both kinds of knowledge are useful when we solve a particular task.
Taking an example for interpretation,
Chinese question answering can take advantage of both cross-task knowledge from Chinese named entity recognition, like the entity information or the boundary information, and cross-language knowledge from English QA, like the linguistic reasoning abilities.

In this paper, we explore to train a general model that enables knowledge to be interacted and shared across tasks and languages simultaneously.
To achieve this goal, we equip our model with trainable and reusable language-specific skill modules and task-specific skill modules.
When the model is applied to a task, it only sparsely activates a subset of the skill modules which are relevant to the target task.
By this way, different tasks in different languages are able to exchange knowledge through skill modules.
In turn, skill modules can also learn corresponding skills.
We can also pre-train the skill modules before multilingual multitask training to get better initialized parameters.
Furthermore, the model can be quickly adapted to new tasks by combining the relevant skill modules learned from other tasks and languages.

Specifically, we propose a multilingual multitask learning model named SkillNet-X.
It is developed with the multi-layer Transformer \cite{attentionVaswaniSPUJGKP17}, which is widely used in different natural language processing tasks. 
We explore two ways to accommodate skill modules by modifying the feed forward network (FFN) layer and the multi-head attention (MHA) layer. 
We evaluate SkillNet-X on eleven datasets from diverse natural language understanding tasks in four languages, including English, Chinese, German and Spanish.
To further investigate the generalization of SkillNet-X, we conduct experiments on two new tasks.
From our experiment results, we have these key findings:
\begin{itemize}
    \item With a general model, our approach performs better than task-specific fine-tuning and two multi-task learning baselines
    on 11 tasks in 4 languages.
    \item Sparsely activated skill pre-training significantly improves the overall performance.
    \item SkillNet-X converges faster and works better when adapted to new tasks by combining the learned skills from other tasks and languages. 
\end{itemize}

\begin{figure*}[!t]
	\centering
	\subfigure[SkillNet-X\textsubscript{FFN}]{
		\includegraphics[width=0.47\linewidth]{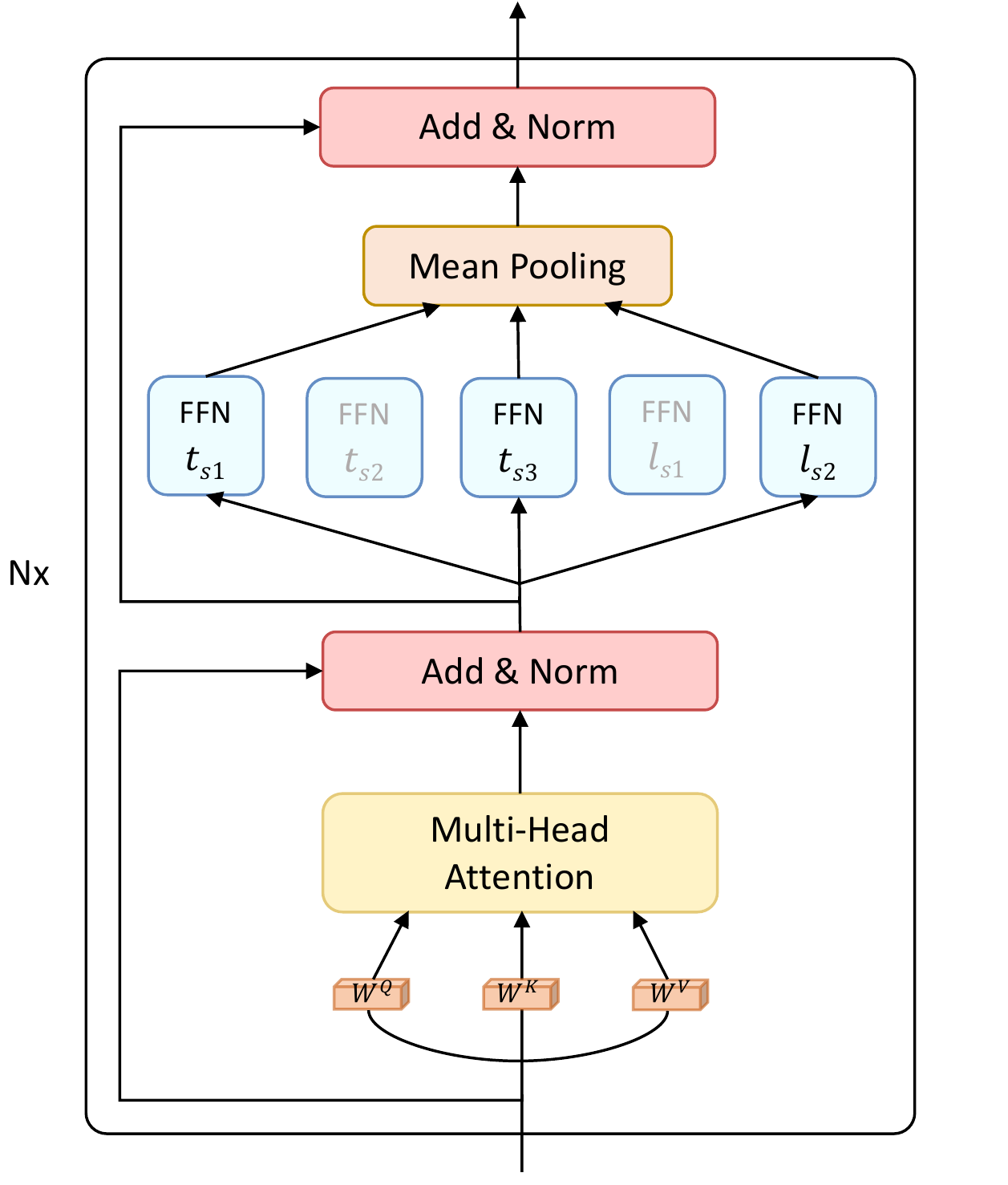}
		\label{fig:model_SkillNet-X_ffn}
	}
	\quad
	\subfigure[SkillNet-X\textsubscript{FFN-MHA}]{
		\includegraphics[width=0.47\linewidth]{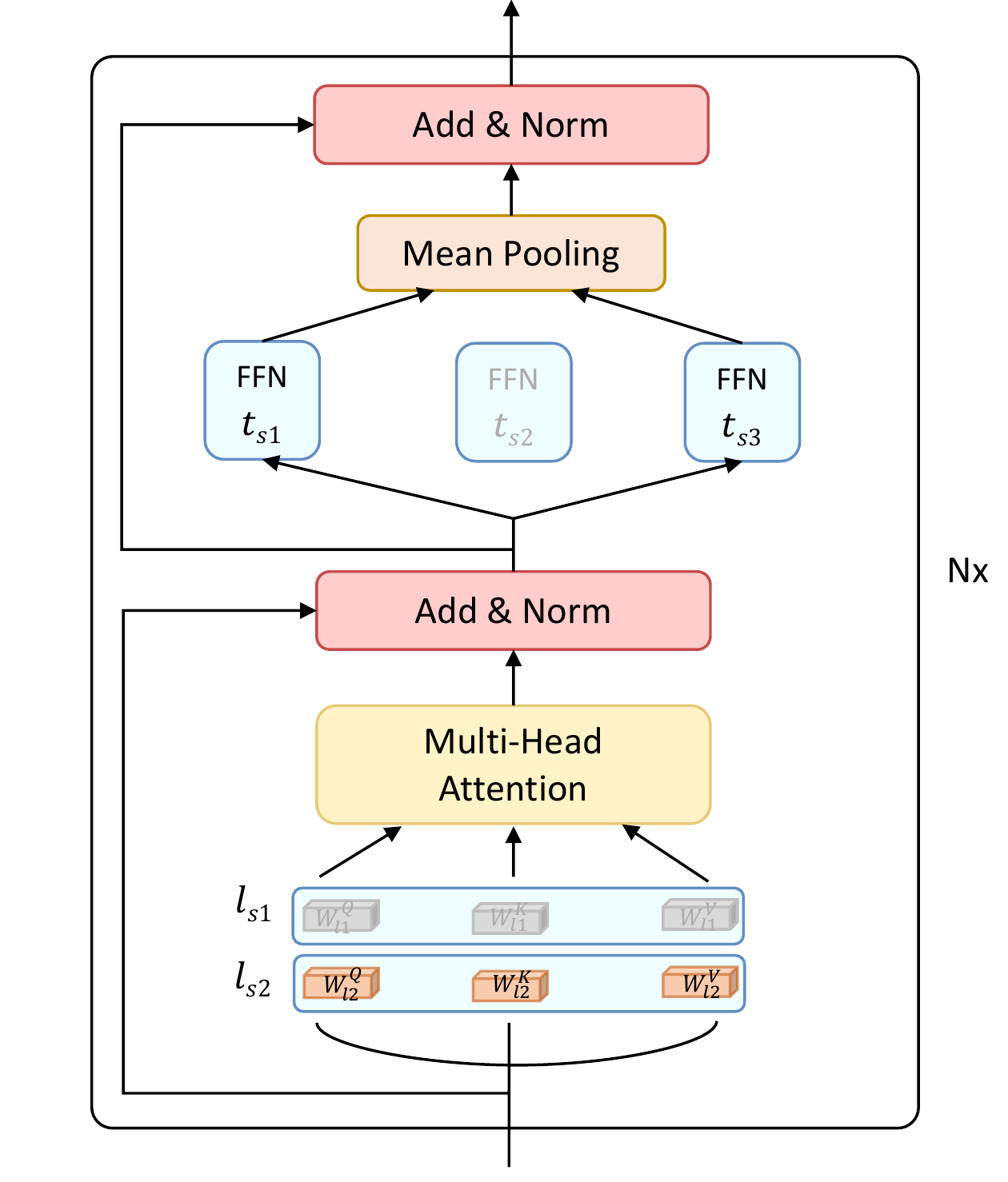}
		\label{fig:model_SkillNet-X}
		} 
	\caption{Architecture of SkillNet-X\textsubscript{FFN} and SkillNet-X\textsubscript{FFN-MHA}. There are 3 task skills (i.e., $t_{s1}$, $t_{s2}$, $t_{s3}$) and 2 language skills (i.e., $l_{s1}$, $l_{s2}$), of which
	$t_{s1}$, $t_{s3}$ and $l_{s2}$ are activated. (a) SkillNet-X\textsubscript{FFN}: all skills are designed and distributed in a sparse Skill FFN layer. 
	(b) SkillNet-X\textsubscript{FFN-MHA}: task skills and language skills are allocated in sparse FFN layer and sparse MHA layer, respectively. Specifically, each group of linear projection parameters stand for a language skill (e.g., the parameters for language skill $l_{s2}$ are $W_{l2}^{Q}$, $W_{l2}^{K}$ and $W_{l2}^{V}$).
	}
	\label{fig:two_models}
\end{figure*}

\section{Definition of Skills}
\label{label:skills}
In order to handle different tasks in different languages, we define task skills  and language skills.
Task skills are used to deal with a specific natural language understanding task.
Language skills are used to understand the basic semantics of a linguistic expression.
Task skills and language skills are complementary.
Both types of skills are required to solve a certain task in a specific language.

As shown in Table~\ref{tab:skills-definition}, in this work, we define a total of ten skills, including six task skills and four language skills.
For a downstream task, we only activate a subset of skills.
For example, to solve the task of sentiment analysis in English, we use two task skills ($t_{s2}$: get the semantic representation of a sentence, and $t_{s5}$: understand the sentiment of text) and one language skill ($l_{s1}$: understand English texts).
Therefore, only the skill parameters corresponding to $t_{s2}$, $t_{s5}$ and $l_{s1}$ are activated.

It should be noted that such a skill set is set for our test datasets according to our prior knowledge. The focus of this paper is to verify the effectiveness of the overall framework. Maybe there's a better setup that we'll explore in the future.

\begin{table}[!ht]
\centering
\begin{tabular}{ll}
\toprule
\bf Skill & \bf Description \\
\midrule
\multicolumn{2}{l}{\it Task Skills} \\
\midrule
$t_{s1}$ & generic skill \\  
$t_{s2}$ & get the semantic meaning of a token \\ 
$t_{s3}$ &  get the semantic meaning of a sentence \\  
$t_{s4}$ &  understand how two text segments interact\\  
$t_{s5}$ &  understand the sentiment of text\\ 
$t_{s6}$  &  understand natural language questions\\  
\midrule
\multicolumn{2}{l}{\it Language Skills} \\
\midrule
$l_{s1}$ & understand English texts \\
$l_{s2}$ & understand Chinese texts \\
$l_{s3}$ & understand German texts \\
$l_{s4}$ & understand Spanish texts \\
\bottomrule
\end{tabular}
\caption{An overview of skills.}
\label{tab:skills-definition}
\end{table}
\section{Approach}
\label{label:model}
In this section, we describe the details about our approach.
We first briefly introduce  background on Transformer (\S \ref{back:transformer}).
Afterwards, we present two ways to accommodate skill modules and explore two variants of SkillNet-X: SkillNet-X\textsubscript{FFN} (\S \ref{model_1}) and SkillNet-X\textsubscript{FFN-MHA} (\S \ref{model_2}).
Finally, we introduce the training process (\S \ref{model_training}) and how to fine-tune SkillNet-X when it is applied to new tasks (\S \ref{model_adaptation}).
\subsection{Background on Transformer}
\label{back:transformer}
Transformer \cite{attentionVaswaniSPUJGKP17} is a widely used model architecture with multiple layers.
Each layer has two important sub-layers: a multi-head attention sub-layer and a feed forward network sub-layer.

Multi-head attention mechanism concatenates the representations of all heads,  and then performs a linearly projection to obtain the final representation.
In each head, $Q$, $K$ and $V$ are linearly projected with different parameter matrices $W_i^{Q}$, $W_i^{K}$ and $W_i^{V}$ as follows:
\begin{equation}
    \mathrm{head_i}= \mathrm{Attention}(QW_i^{Q} , KW_i^{K}, VW_i^{V}).
\end{equation}

In addition to multi-head attention, each of the layers contains a fully connected feed forward network.
This consists of two linear transformations with a ReLU activation in between as follows:
\begin{equation}
\label{e2}
\mathrm{FFN}(x) = \max(0, xW_1 + b_1) W_2 + b_2,
\end{equation}
where $W_1$, $b_1$, $W_2$ and $b_2$ are learned parameters, $x$ is an input token representation.

\subsection{Model \uppercase\expandafter{\romannumeral1}: SkillNet-X\textsubscript{FFN}}
\label{model_1}
As shown in Figure~\ref{fig:model_SkillNet-X_ffn}, we modify the dense feed forward network (FFN) of each Transformer layer with a sparse Skill FFN layer.

Specifically, we have a set of FFN layers in parallel, in which each FFN layer stands for a task-specific skill or language-specific skill.
Instead of producing general representations for all tasks in FFN layer, we activate different skill modules parameters to produce different skill-specific representations.
When the model is applied to a task, unlike traditional dense models that always activate all the model parameters, it sparsely activates parts of the parameters whose skills are relevant to the target task.
What skills are activated by a task is guided by a manually defined task-skill binary matrix.
Multiple task-specific skills can be activated simultaneously, while only one language-specific skill can be activated at a time.
We activate the same skill modules for the same task.

Given an input token $x$, we get a set of skill-specific representations $S$ through Equation \ref{e3}.
Considering that the number of skill modules activated by different tasks may be different, we adopt mean-pooling to obtain the final representation.
We give the formalized description as follows:
\begin{gather}
S =  \begin{bmatrix}
\mathrm{FFN}_1(x) & \mathrm{FFN}_2(x)  & \cdots & \mathrm{FFN}_n(x) 
\label{e3}
\end{bmatrix}, \\ 
\mathrm{FFN^{Skill}}(x) =  \mathrm{MeanPooling}(M_{(x)} \odot S), \\
M = \begin{bmatrix}
1 & 0 & \cdots & 1 \\
0 & 1 & \cdots & 1\\
1 & 1 & \cdots & 0 \\
\vdots & \vdots & \ddots & \vdots \\
0 & 1 & \cdots & 0 \\
\end{bmatrix},
\end{gather}
where $n$ is the total number of skills, $S$ stands for $n$ skill-specific representations, $M$ is the task-skill binary matrix.
$M_{(x)}$ is a row of $M$, which indicates the skills to be activated for the task corresponding to the current input $x$.
In task-skill binary Matrix M, 1 indicates that the skill is activated, while 0 indicates that the skill is deactivated.

\subsection{Model \uppercase\expandafter{\romannumeral2}: SkillNet-X\textsubscript{FFN-MHA}}
\label{model_2}
As shown in Figure~\ref{fig:model_SkillNet-X}, we modify the dense FFN layer and the multi-head attention (MHA) layer.
The FFN layer is modified similar with SkillNet-X\textsubscript{FFN}.
The multi-head attention layer is replaced with a sparse multi-head attention layer.

Specifically, we modify the multi-head attention of each Transformer layer as follows.
We have a set of attention modules in parallel, each of which stands for a language-specific skill.
Instead of producing general $Q/K/V$ vectors for all languages, we activate different skill parameters to produce different language-specific $Q/K/V$ vectors before conducting multi-head attention.
Take $Q$ as an example.
Instead of having only one projection matrix for all queries, we have four projection parameter matrices $\{W_{l1}^{Q}, W_{l2}^{Q}, W_{l3}^{Q}, W_{l4}^{Q} \}$, of which each item stands for a skill of understanding the information of a particular language.
When the model is applied to a task, we only activate the corresponding  projection matrices.
Similar modifications are made for keys and values. 

Suppose we have an input token $x$, which comes from a Chinese task.
The computation of a head is modified as follows.
\begin{equation}
   \mathrm{head_i^{Skill}}(x) = \mathrm{Attention}\small(Q{{W}_{l2}^{Q}}_i, K{{W}_{l2}^{K}}_i, V{{W}_{l2}^{V}}_i),
\end{equation}
where ${{W}_{l2}^{Q}}_i$, ${{W}_{l2}^{K}}_i$ and ${{W}_{l2}^{V}}_i$ are the parameters for language skill $l_{s2}$.

\begin{table*}[t]
\centering
\begin{tabular}{ccccccccccccc}
\toprule
\multirow{2}[3]{*}{\bf Dataset Id} & \multirow{2}[3]{*}{\bf Dataset} & \multirow{2}[3]{*}{\bf Language} & \multicolumn{6}{c}{\bf Task Skills} & \multicolumn{4}{c}{\bf Language Skills} \\
\cmidrule(lr){4-9} \cmidrule(lr){10-13}
& &  & $t_{s1}$ & $t_{s2}$ & $t_{s3}$ & $t_{s4}$ & $t_{s5}$ & $t_{s6}$ & $l_{s1}$ &  $l_{s2}$ & $l_{t3}$ & $l_{t4}$ \\
\midrule
\text{T1} & MNLI &  en & \checkmark & & \checkmark & \checkmark &  & & \checkmark & & & \\
\text{T2} & QNLI & en &  \checkmark & & \checkmark & \checkmark & & \checkmark &\checkmark &   & &  \\
\text{T3} & SST-2 & en &  \checkmark & & \checkmark & & \checkmark  & & \checkmark &  & &  \\
\text{T4} & SQuAD 1.1 & en &   \checkmark & \checkmark & & \checkmark & & \checkmark & \checkmark&  & &  \\
\text{T5} & CoNLL03 & en  & \checkmark & \checkmark & & & & & \checkmark&  &  & \\
\text{T6} & OCNLI & zh  &  \checkmark & & \checkmark & & &  & & \checkmark & & \\
\text{T7} & TNEWS & zh  & \checkmark & & \checkmark & & & &  & \checkmark& & \\
\text{T8} & MARC-de & de  & \checkmark &  & \checkmark &&&&&& \checkmark &  \\
\text{T9} & Wikiann-de &  de & \checkmark & \checkmark &&&&&&& \checkmark &  \\
\text{T10} & MARC-es & es  & \checkmark & &  \checkmark &&&&&&& \checkmark \\
\text{T11} & SQuAD-es & es & \checkmark & \checkmark & & \checkmark & & \checkmark &&&& \checkmark \\
\bottomrule
\end{tabular}
\caption{Details about eleven datasets used to train SkillNet-X. Relevant skills (defined in Table \ref{tab:skills-definition}) for each dataset are marked with a tick. ``en'', ``zh'', ``de'' and ``es'' denote English, Chinese, German and Spanish languages.}
\label{tab:task-skill-map}
\end{table*}

\subsection{Model Training}\label{model_training}
\label{model_training}
The training objective is joint minimizing the sum of loss functions on different tasks in different languages.
Following \cite{Zhang2022SkillNetNLUAS}, the model is trained on the concatenation of training samples from these tasks.
In each iteration, a mini-batch is selected from one task, and the model parameters are updated according to the task-specific objective.

\paragraph{Data Sampling}
The large difference of data amount among different datasets will lead to the model bias towards the dataset with large data amount.
To alleviate this problem, we sample mini-batches from all tasks according to a multinomial distribution as  \cite{xlm-ConneauL19}, with probabilities $q_i$:
\begin{equation}
q_{i} = \frac{p_{i}^{\alpha}}{\sum_{j=1}^{N} p_{j}^{\alpha}}\ \ \text{with} \ \ p_{i} = \frac{|T_{i}|}{\sum_{k=1}^{N}|T_{k}|},
\end{equation}
where $|T_{i}|$ indicates the number of training examples in task $T_i$.
The sampling rate $\alpha$ is a hyper-parameter to balance various tasks.

\paragraph{Loss scaling}
In multitask learning, the number of classes of different tasks is different.
To ensure stability in the training process, we leverage loss scaling to normalize the task-specific loss function with respect to the number of classes in the tasks as follows:
\begin{equation}
    \mathcal{L} = \sum_{t}^{N} \frac{\mathcal{L}_{t} }{log{C_t}},
\end{equation}
where $N$ is the number of tasks, $C_t$ denotes the number of classes in task $t$,  $\mathcal{L}_{t}$ is the loss of task $t$.
\paragraph{Skill Pre-training}
In this section, we introduce how the skill module can be pre-trained with model parameters being sparsely activated.
We adopt two standard self-supervised learning objectives \cite{bert} including masked language modeling (MLM) and next sentence prediction (NSP) for each language.
We activate two task skills $\{t_{s1}, t_{s2}\}$ for the MLM task and three task skills $\{t_{s1}, t_{s3}, t_{s4}\}$  for the NSP task. 
To avoid introducing extra data, we use English, Chinese, German and Spanish parts of XLM-R \cite{xlm-roberta-ConneauKGCWGGOZ20} pre-training data for skill pre-training.
The parameters of pre-trained skills can be used to initialize the model after being pre-trained.

\subsection{Adaptation to New tasks}
\label{model_adaptation}
We can combine the well-trained skill modules to solve new tasks, which are never seen in the multilingual multitask training stage.
For example, although the model has not been trained on Spanish named entity recognition (NER) task,  we can use the relevant skills (i.e., the skill of getting the semantic of a token ($t_{s2}$) and the skill of understanding Spanish texts ($l_{s4}$)) to solve this task.
Details are given in the experiment section.

\section{Experiments}
This section is organized as follows.
We first describe the datasets for multilingual multitask training, and then describe the experimental settings.
\subsection{Datasets}
As shown in Table~\ref{tab:task-skill-map} and Table~\ref{tab:data_size}, we use eleven datasets from four languages, including English, Chinese, German and Spanish, to train our models .
Data statistics of the training/validation data splits for eleven datasets are given in Table~\ref{tab:data_size}. We will introduce each dataset in detail in the following.

\paragraph{English}
We use five datasets, including MNLI, QNLI, SST-2, SQuAD 1.1 and CoNLL03.
MNLI, QNLI and SST-2 are from English GLUE benchmark \cite{wang-etal-2018-glue}.
MNLI \cite{williams-etal-2018-broad} is a large-scale multi-genre natural language inference datasets.
Given a pair of sentences, the task is to predict the
relationship between two sentences: \textit{entailment}, \textit{contradiction}, or \textit{neutral}. In this work, we use the matched part of MNLI.
QNLI is a variant of the SQuAD dataset \cite{rajpurkar-etal-2016-squad}.
The task is to determine whether the context sentence contains the answer to the question.
SST-2 is the Stanford Sentiment Treebank \cite{socher-etal-2013-recursive}, which is a binary single-sentence classification task.
SQuAD 1.1 is the Stanford Question Answering Dataset \cite{rajpurkar-etal-2016-squad}, which provides a paragraph of context and a question.
The task is to answer the question by extracting relevant span from the context.
CoNLL03 \cite{tjong-kim-sang-de-meulder-2003-introduction} is a widely use named entity recognition dataset released as a part of CoNLL-2003 shared task.
\begin{table}[!ht]
\centering
\begin{tabular}{lccc}
\toprule
\bf Dataset & Language & \#Train & \#Dev  \\
\midrule
MNLI & en & 393k & 9.8k   \\
QNLI & en & 105k & 5.5k  \\
SST-2 & en & 67.3k & 872\\
SQuAD 1.1 & en & 11.5k & 2.2k  \\
CoNLL03 & en & 14k & 3.3k \\
OCNLI & zh & 50k & 3k \\
TNEWS & zh & 53k & 10k \\
MARC-de & de &  200k & 5k \\
Wikiann-de & de & 20k & 10k  \\
MARC-es & es & 200k & 5k \\
SQuAD-es & es & 88k & 10.6k\\
\bottomrule
\end{tabular}
\caption{Details about eleven datasets used to train SkillNet-X.}
\label{tab:data_size}
\end{table}
\paragraph{Chinese}
We use TNEWS dataset and OCNLI dataset, which are both from Chinese CLUE benchmark \cite{xu-etal-2020-clue}.
TNEWS \cite{xu-etal-2020-clue} is a short text classification dataset for news titles. 
Each title is labeled with one of 15 news categories (sports, education, game, etc.).  
OCNLI \cite{hu-etal-2020-ocnli} is an original Chinese natural language inference dataset.
The premises are collected from Chinese sources and the hypotheses are written by human.
Similar with MNLI, the task is to predict the relation between premise and hypothesis.

\paragraph{German}
\label{dataset:german}
We use MARC-de dataset and Wikiann-de dataset.
MARC \cite{keung2020multilingual} is the Multilingual Amazon Reviews Corpus, which contains reviews in several languages.
Given a review, the task is to predict the star rating on a 5-star scale.
Wikiann \cite{pan-etal-2017-cross} is a cross-lingual NER corpus.
We adopt the same train, dev, and test splits with XTREME \cite{hu2020xtreme}.
MARC-de and Wikiann-de means the German part of MARC and Wikiann, respectively.

\begin{table*}[t]
\centering
\small
\begin{tabular}{lcccccccccccc}
\toprule
Model & \text{T1} & \text{T2} & \text{T3} & \text{T4} & \text{T5} & \text{T6} & \text{T7} & \text{T8} & \text{T9} & \text{T10} & \text{T11} & \bf Avg. \\
\midrule
Metrics & Acc. & Acc. & Acc. & F1 & F1 & Acc. & Acc. & Acc. & F1 & Acc. & F1 &  \\
\midrule
Task-specific Fine-tuning & \bf 84.4 &	90.2 &	92.8 &	88.3 	&94.3 &	73.9 &	55.7 	&62.9 &	86.2 &	59.1 &	76.2 &	78.5   \\
Joint Fine-tuning (Dense) & 82.7 &	91.5 & 	\bf 93.5 &	89.0 &	94.7 &	75.5 &	55.9 &	63.7 &	87.0 &	59.4 &	76.5 &	79.0    \\
Joint Fine-tuning (MoE)& 81.9 & 91.3 & 92.2 & 88.9 & 95.0 & 75.5 & 56.2 & 63.2 & 88.5 & 59.3 & 76.7 & 79.0 \\
\midrule
SkillNet-X\textsubscript{FFN} w/o SP  & 83.7 &	91.5 &	92.9 &	89.0 &	95.3 	&76.3 	&56.1 &	63.1 &	88.8 &	59.2 &	76.8 	&79.3 \\
SkillNet-X\textsubscript{FFN-MHA} w/o SP & 84.2 & 91.4 & \bf 93.5 & 89.1 & 95.2 & \bf 76.8 & 55.3 & 63.9 & 88.2 & 59.4 & 77.1 & 79.4 \\
SkillNet-X\textsubscript{FFN} & 83.5 &	91.6 &	92.8 &	89.8 &	95.3 &	76.5 &	56.5 &	63.6 &	\bf 89.3 	& 59.6 & 	77.8 &	79.6  \\
SkillNet-X\textsubscript{FFN-MHA} & 84.1 &  \bf 91.8 & 92.9 &  \bf 90.5 & \bf 95.6 & 76.3 & \bf 57.2 & \bf 64.2 &  \bf 89.3 &  \bf 60.0 & \bf 78.3 &  \bf 80.0 \\
\bottomrule
\end{tabular}
\caption{Results on eleven datasets. \text{Avg.} is the average score of all datasets. \textbf{Bold} text denotes the best result in each column. Results with $^{\dagger}$  are from \cite{bert}. Results with $^{\ddagger}$ are from \cite{xu-etal-2020-clue}. w/o SP means that skills are not pre-trained.}
\label{tab:main_results}
\end{table*}

\begin{table*}[t]
\centering
\begin{tabular}{lccccccc}
\toprule
 & MNLI & QNLI & SST-2 & SQuAD 1.1 & CoNLL03  & Avg. & $\Delta$ \\
\midrule
SkillNet-X\textsubscript{FFN-MHA} & 84.1 &  91.8 & 92.9 & 90.5 & 95.6 & 91.0 & - \\
\midrule
\multicolumn{5}{l}{\it  Language Skill Perturbation} \\ 
\midrule
en $\rightarrow$ \text{zh} & 78.2 & 84.1 & 90.7 & 83.3 & 92.2 & 85.7 & -5.3\\
en $\rightarrow$ \text{de} & 78.6 & 85.6 & 90.5 & 82.6 & 64.5 & 80.4 & -10.6\\
en $\rightarrow$ \text{es} & 77.7 & 90.1 & 92.2 & 87.4 & 92.9  & 88.1 & -2.9\\
\midrule
\multicolumn{5}{l}{\it Task Skill Perturbation} \\
\midrule
All Skills & 72.5 & 88.5 & 89.3 & 86.6 & 42.1 & 75.8 &  -15.2 \\
Random Skills (seed=42) & 40.5 & 	81.0& 	57.1 &	75.8 &	26.9 &	56.3  & -34.7 \\
Random Skills (seed=624) & 77.8 & 67.8 &	69.8 &	81.6 &	3.5 &	60.1 & -30.9 \\
Random Skills (seed=123) & 38.5 &	83.1 & 	88.5 & 	81.4 &	48.8 & 	68.1 & -22.9  \\
\bottomrule
\end{tabular}
\caption{Results on five English datasets with skill perturbation. Based on original model (first group), we add language skill perturbation (second group) and task skill perturbation (third group) during inference.  en $\rightarrow$ zh means that we replace English language skill with Chinese language skill. Avg. is the macro-average score of five datasets. $\Delta$ denotes the change of average score.}
\label{tab:skill-analysis-results}
\end{table*}

\paragraph{Spanish}
We use MARC-es dataset and SQuAD-es 1.1 dataset.
MARC-es \cite{keung2020multilingual} stands for the Spanish part of MARC, similar with MARC-de.
SQuAD-es 1.1 \cite{carrino-etal-2020-automatic-essquad} is a large-scale question answering training resource for Spanish, which is an automatically translated version of English SQuAD 1.1.

\subsection{Experimental Setup}
\paragraph{Baselines}
We compare our SkillNet-X with the following three approaches:
\begin{itemize}
	\item \textbf{Task-specific Fine-tuning}: Following the standard task-specific fine-tuning, We fine-tune all the parameters of the model for each task individually. As a result, we get eleven task-specific models in total. 
	\item \textbf{Joint Fine-tuning (Dense)}: Following MT-DNN \cite{liu-etal-2019-multi}, we use the dense joint training method in multi-task learning to jointly fine-tune the model on the eleven datasets. All datasets share the parameters of the encoder.
	\item \textbf{Joint Fine-tuning (MoE)}: Following \cite{shazeer2017outrageously}, we replace the FFN with the Mixture-of-Experts (MoE) architecture containing $n$ experts in each Transformer block. The sparse gating network activates the top-2 experts for each token. To ensure a fair comparison, we maintain a consistent number ($n=10$) of experts in this model and skills in SkillNet-X.
\end{itemize}
\paragraph{Implementation Details}
All baselines and our models are initialized with pre-trained XLM-R \cite{xlm-roberta-ConneauKGCWGGOZ20} \footnote{\href{https://huggingface.co/xlm-roberta-base}{https://huggingface.co/xlm-roberta-base}}.
For task-specific fine-tuning setting, we set max epochs to 3 for each dataset.
For other multitask fine-tuning settings, we set max training steps to 200K \footnote{In both cases,  models are trained with roughly the same amount of training data.}.
We set learning rate to 1e-5, set batch size to 16 and use Adam optimizer \cite{kingma2014adam} with linear learning rate decay schedule to update the parameters.
We set sample factor $\alpha$ to 0.6 for dense model and MoE model, and set $\alpha$ to 0.4 for our models.
We utilize PyTorch and HuggingFace Transformers for our implementation.

\section{Results and Analysis}

\subsection{Multitask Results}
Results are shown in Table~\ref{tab:main_results}.
We report results on each dataset and the macro-average score on eleven datasets.
Two multitask learning baselines (i.e., Joint Fine-tuning (Dense) and Joint Fine-tuning (MoE)) perform slightly better than task-specific fine-tuning.
Even without skill pre-training, our models (SkillNet-X\textsubscript{FFN} w/o SP and SkillNet-X\textsubscript{FFN-MHA} w/o SP) can achieve consistent and remarkable improvements over all baselines on almost all datasets, demonstrating its effectiveness.
By pre-training the skills, both SkillNet-X\textsubscript{FFN} and SkillNet-X\textsubscript{FFN-MHA} further boost the performances by a large margin (i.e., 79.3 $\rightarrow$ 79.6 and 79.4 $\rightarrow$ 80.0).
And SkillNet-X\textsubscript{FFN-MHA} achieves the highest average score.

\begin{table*}[t]
\centering
\begin{tabular}{lcccc}
\toprule
& SkillNet-X\textsubscript{FFN-MHA} & \makecell[c]{Random Skills  \\ (seed=42) } & \makecell[c]{Random Skills \\ (seed=624)} & \makecell[c]{Random Skills \\ (seed=123)} \\
\midrule
MNLI&  $t_{s1}$,$t_{s3}$,$t_{s4}$ & $t_{s1}$,$t_{s5}$,$t_{s6}$ & $t_{s4}$ & $t_{s5}$ \\
QNLI & $t_{s1}$,$t_{s3}$,$t_{s4}$,$t_{s6}$&  $t_{s1}$,$t_{s6}$ & $t_{s1}$,$t_{s2}$,$t_{s3}$,$t_{s5}$,$t_{s6}$ & $t_{s1}$,$t_{s3}$ \\
SST-2 & $t_{s1}$,$t_{s3}$,$t_{s5}$ & $t_{s3}$,$t_{s4}$,$t_{s6}$ & $t_{s1}$,$t_{s2}$,$t_{s3}$,$t_{s4}$,$t_{s6}$ & $t_{s5}$\\
SQuAD 1.1 & $t_{s1}$,$t_{s2}$,$t_{s4}$,$t_{s6}$ & $t_{s1}$,$t_{s3}$,$t_{s4}$ & $t_{s1}$,$t_{s2}$,$t_{s3}$,$t_{s5}$,$t_{s6}$ & $t_{s3}$,$t_{s6}$\\
CoNLL03 & $t_{s1}$,$t_{s2}$ & $t_{s1}$,$t_{s5}$,$t_{s6}$ & $t_{s3}$,$t_{s5}$ & $t_{s2}$,$t_{s3}$,$t_{s5}$,$t_{s6}$ \\
\bottomrule
\end{tabular}
\caption{Activated skills with different random seeds for each English dataset.}
\label{tab:skill_activate}
\end{table*}

\subsection{Analysis of Skill Module}
\label{skill_perturbation}
To verify whether skill modules have learned the corresponding skills, we add some skill perturbations and then observe how the model performances change.
Specifically, in the inference stage, we add language skill perturbation and task skill perturbation, respectively.
Based on the model SkillNet-X\textsubscript{FFN-MHA}, we conduct experiments on the English datasets.
For language skill perturbation, we employ Chinese, German or Spanish language skill instead of the original English language skill.
For task skill perturbation, instead of activating the corresponding task skills defined in Table~\ref{tab:task-skill-map}, we attempt two strategies: (1) all task skills are activated for each task. (2) each skill is activated randomly with a probability of 0.5 for each task.

Table~\ref{tab:skill-analysis-results} shows that both kinds of skill perturbations will affect the effectiveness of the model.
When English language skill is replaced with other language skills, the performance of model drops significantly (e.g., 10.6 points drops in en $\rightarrow$ de).
In terms of task skill perturbation, we can see that the average score drops by 15.2 points ($91.0 \rightarrow 75.8$) when all task skills are used.
If we randomly activate each skill for each dataset, the average score drops even more.
As shown in Table~\ref{tab:skill-analysis-results}, the average scores dropped by 34.7, 30.9 and 22.9 for three different random tests.
To further analyze these results, as shonw in Table~\ref{tab:skill_activate}, we compare sampled skills with original skills for MNLI, QNLI, SST-2, SQuAD 1.1 and CoNLL03 datasets.
We can see that the less difference between sampled task skills and original skills, the higher the average score, which indicates skill modules have learned the corresponding skills.

\subsection{Influence of the Sampling Factor}
\begin{figure}[!ht]
\centering
\includegraphics[width=\linewidth]{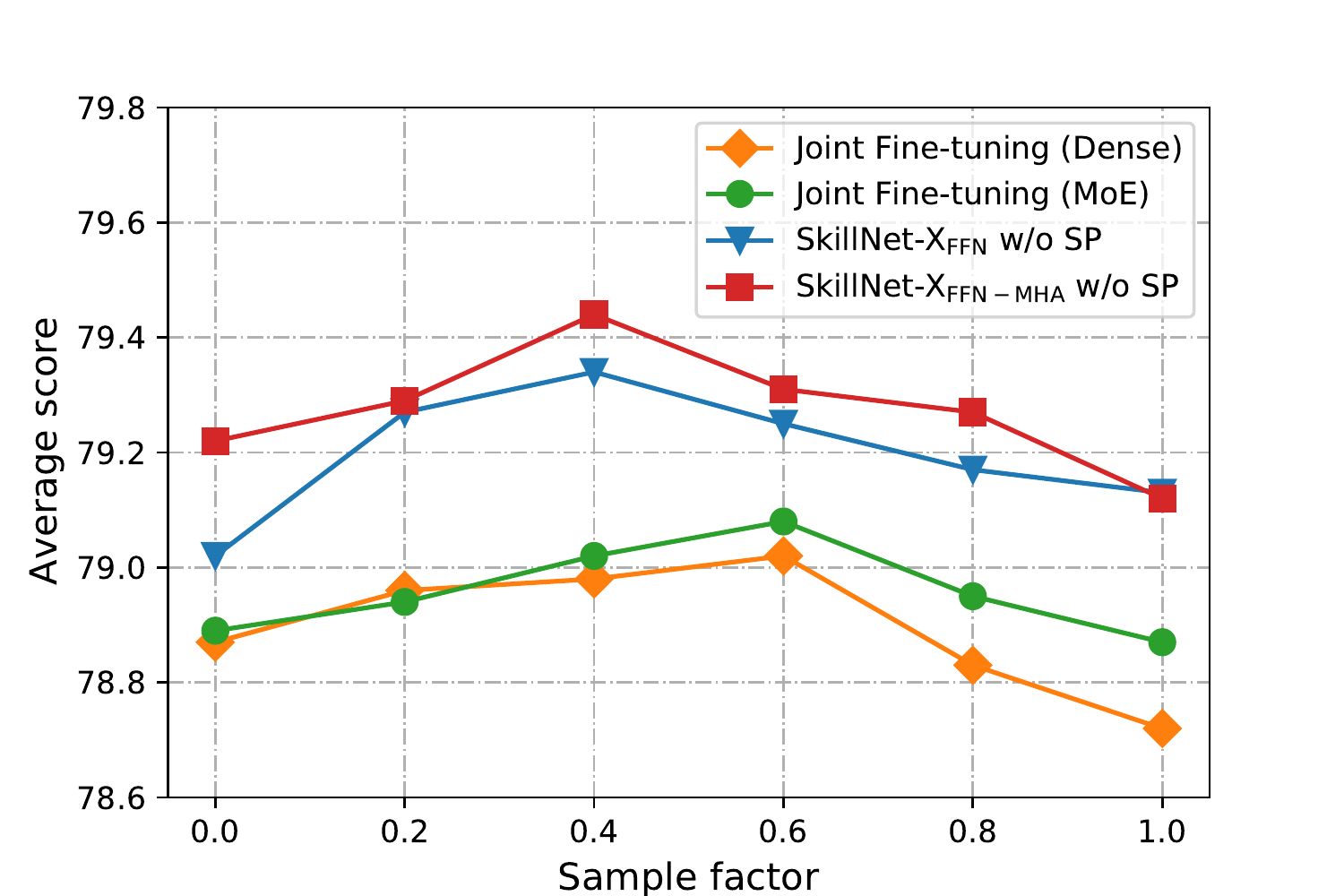}
\caption{Average score with different sample factor $\alpha$.}
\label{fig:sample}
\end{figure}
As introduced in Section \S \ref{model_training}, we sample mini-batch from each task according to the sampling rate $\alpha$.
Figure~\ref{fig:sample} shows the average score with different $\alpha$.
As we can see,  Joint Fine-tuning (Dense) and Joint Fine-tuning (MoE) are both perform best when $\alpha$ is 0.6.
SkillNet-X without skill pre-training performs best when $\alpha$ is 0.4, so we  also set $\alpha$ as 0.4 for SkillNet-X model with skill pre-training.

\begin{figure*}[htbp]
\centering
\subfigure[Performance on MRPC dataset with different data sizes.]{
\includegraphics[width=0.47\linewidth]{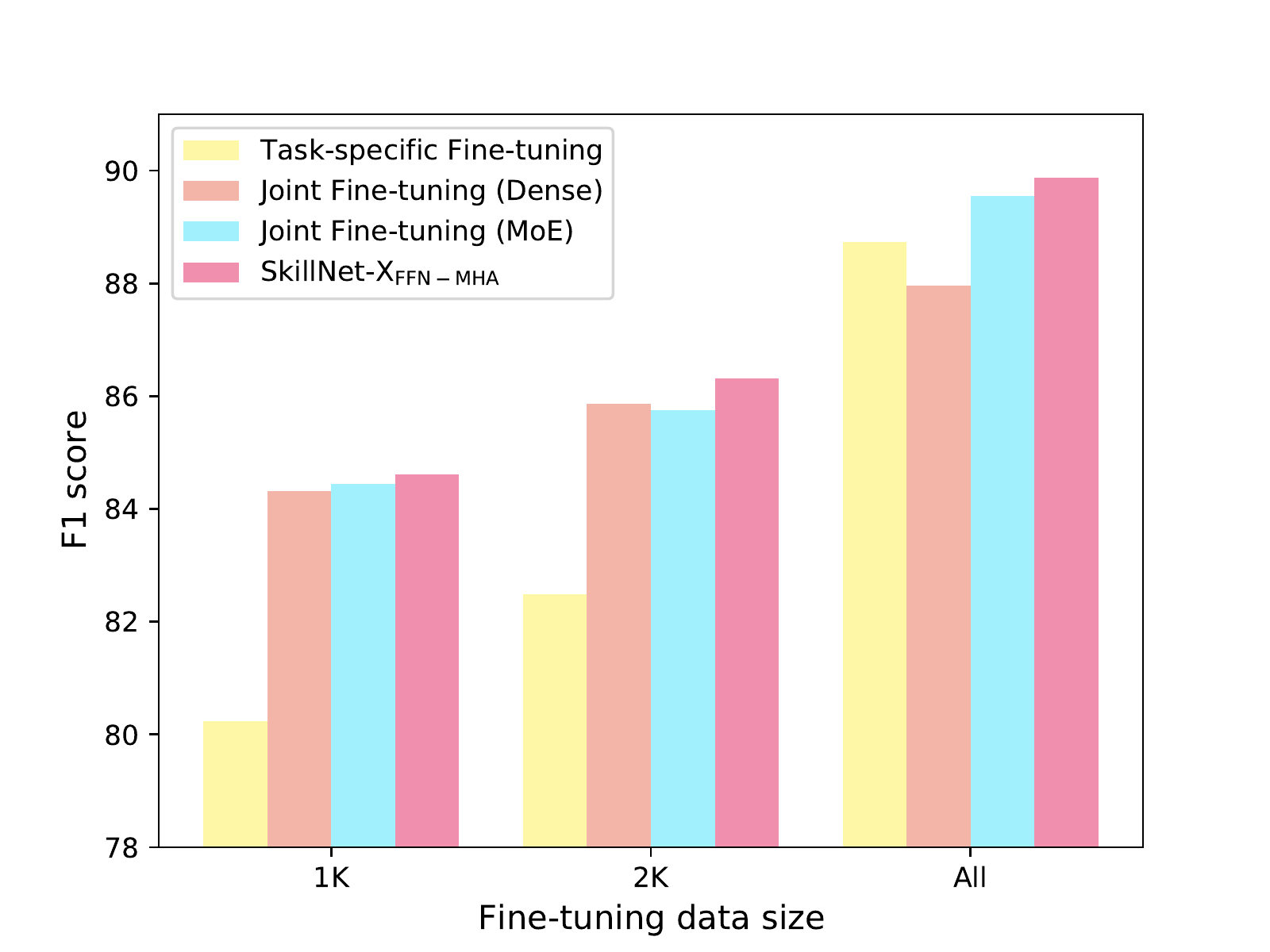}
\label{fig:data_num}
}
\quad
\subfigure[Performance on Wikiann-es dataset with different fine-tuning steps.]{
\includegraphics[width=0.47\linewidth]{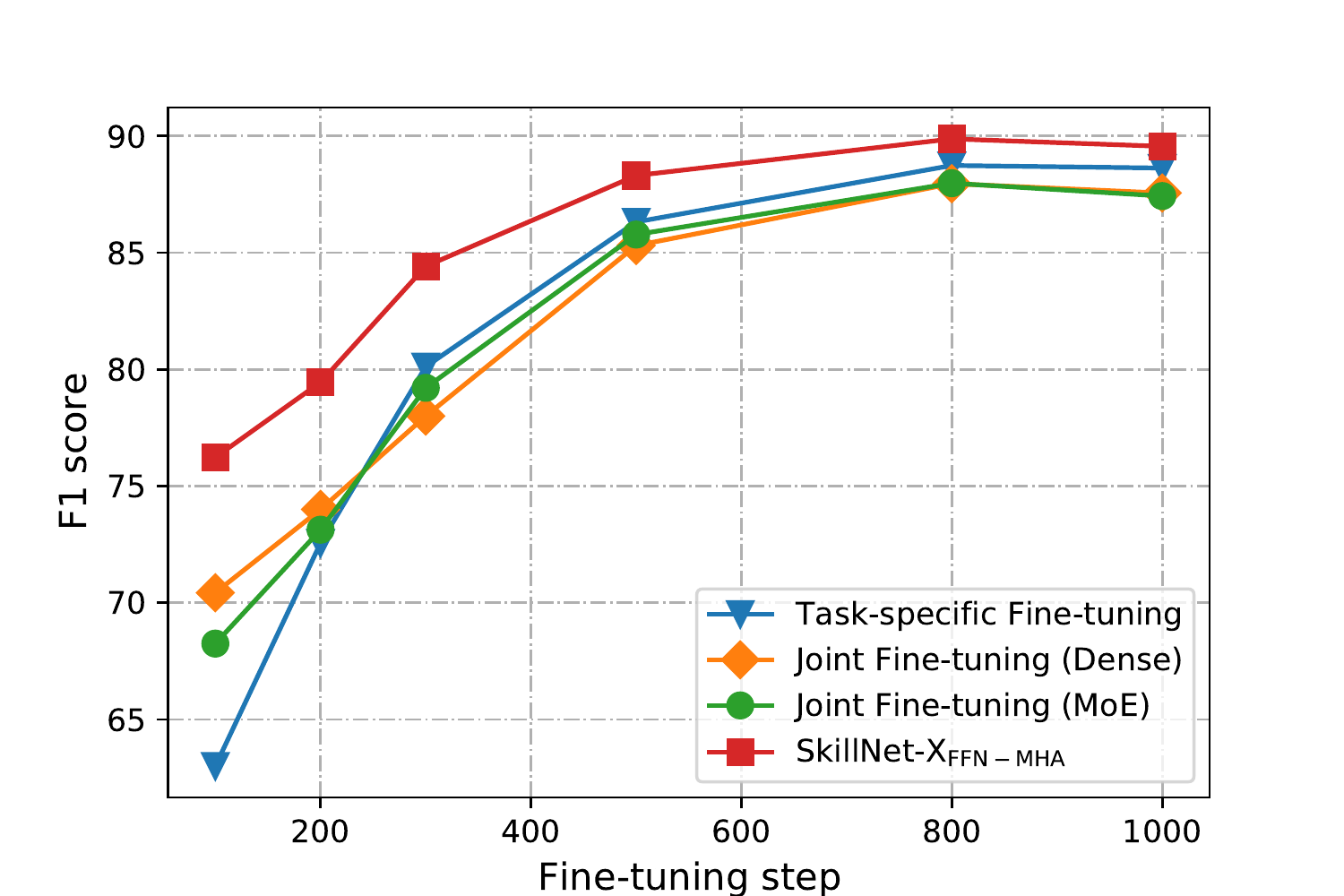}
\label{fig:steps}
}
\caption{Analyze the effect of (a) dataset size and (b) number of fine-tuning steps .}
\end{figure*}

\subsection{New Tasks}
\begin{table}[!ht]
\centering
\begin{tabular}{lccc}
\toprule
\bf Model & MRPC & Wikiann-es \\
\midrule
Task-specific Fine-tuning & 90.62 & 88.74  \\
Joint Fine-tuning (Dense) & 91.55 & 87.96 \\
Joint Fine-tuning (MoE) & 91.54 & 87.98 \\
SkillNet-X\textsubscript{FFN-MHA}  &\bf  92.69 &  \bf 89.88 \\
\bottomrule
\end{tabular}
\caption{Results on MRPC and Wikiann-es datasets. We report the F1 score for both task.}
\label{tab:newtask-results}
\end{table}

We would like to evaluate SkillNet-X\textsubscript{FFN-MHA} on new tasks which are never seen in the multitask training stage.
To this end, we study two new tasks: paraphrase sentences pair classification in English and NER in Spanish.
We conduct experiments on the datasets of MRPC \cite{dolan-brockett-2005-automatically} and Wikiann-es.
MRPC is a English paraphrase task from GLUE Benchmark \cite{wang-etal-2018-glue}.
The task is to determine whether the sentences in the pair are semantically equivalent.
Wikiann-es is the Spanish part of Wikiann, which is introduced in \S \ref{dataset:german}.
We use three task skills $\{ t_{s1}, t_{s3}, t_{s4}\}$ and the English language skill $l_{s1}$ for MRPC dataset, and use two task skills $\{ t_{s1}, t_{s2}\}$ and the Spanish language skill $l_s{4}$ for Wikiann-es dataset.
We fine-tune models with the following set of hyper-parameters: batch size is 16, max epoch is 3, and learning rate is in \{1e-5, 2e-5, 5e-5\}.

Table~\ref{tab:newtask-results} shows the results with different models for two new tasks.
As we can see, SkillNet-X\textsubscript{FFN-MHA} significantly outperforms all baselines (e.g., 90.62 vs. 92.69, 88.74 vs. 89.88), which illustrates that our model could generalize better to new task which is never seen in the multitask training step. To better measure the model's ability to be adapted to new tasks, we also analyze the influence of the number of data size and the fine-tuning step.

\paragraph{Influence of the Number of Data Size}

In order to analyze the impact of the number of data size, we used 1K, 2K, and all examples of MRPC dataset to fine-tune models, respectively.
Results are shown in Figure \ref{fig:data_num}.
As we can see, SkillNet-X\textsubscript{FFN-MHA} can exceed baselines  with different data sizes.

\paragraph{Influence of Fine-tuning Step}
In order to analyze the impact of fine-tuning step, we fine-tune models with different steps on Wikiann-es dataset.
Specifically, we fine-tune models with 100, 200, 300, 500, 800 and 1,000 steps, respectively.
Results are shown in Figure \ref{fig:steps}.
We can see that SkillNet-X\textsubscript{FFN-MHA} gives the model a good starting point and leads to better F1 score.

\section{Related Work}
In this section, we introduce the connections and differences of this work to multilingual learning and multitask learning.
\subsection{Multilingual Learning}
Multilingual learning has been applied in different natural language processing areas.
\cite{firat-etal-2016-multi, aharoni-etal-2019-massively} attempted to build a single multilingual neural translation model to translate between multiple languages.
\cite{giannakopoulos-etal-2015-multiling, scialom-etal-2020-mlsum} applied multilingual learning to the field of text summarization.
\cite{rahimi-etal-2019-massively, tedeschi-etal-2021-wikineural-combined} used multilingual learning to improve the performance of named entity recognition.
However, they mainly focused on the same task in different languages.
In this work, we explore to train a general model to solve different natural language understanding tasks in different languages.
\subsection{Multitask Learning}
Multi-task learning \cite{caruana1997multitask}, which aims to model the useful information among tasks, has been studied extensively in natural language processing \cite{ruder2017overview, craw20multitask}.
Dense joint training is the most commonly used approach to multitask learning.
It is generally applied by sharing the encoder layers between all tasks, while keeping several task-specific output layers, such as MT-DNN \cite{liu-etal-2019-multi}, which uses BERT as encoder to perform multitask learning on GLUE tasks \cite{wang-etal-2018-glue}.
However, this approach suffers from optimization conflicts caused by task differences, because all tasks need to use the same set of parameters on shared-bottom layers.

Instead of sharing same model parameters, some recent approaches began to apply Mixture-of-Experts (MoE) \cite{shazeer2017outrageously} to multitask learning, such as MT-TaG \cite{gupta2022sparsely}.
\cite{shazeer2017outrageously} first proposed the Mixture-of-Experts (MoE) layer with a single gating network with Top-$k$ routing. 
Different with MoE, MT-TaG adopts multiple task-aware gating networks to route examples from different tasks to learn task-specific information.
Considering that it is unclear what type of knowledge is learned in each expert and why an expert is activated, \cite{dai2022one,liao2022skillnet} attempt to define what an expert means and heuristically activate the corresponding expert.
\cite{liao2022skillnet} focus on multitask learning in the same language, and \cite{dai2022one} focuses on multimodal learning.
In contrast, we explore to use sparse model to  multilingual multitask learning.

\section{Conclusion}
This paper presents a sparsely activated multilingual multitask model named SkillNet-X.
We demonstrate that one model is able to deal with multiple natural language understanding tasks in multiple languages, and even achieves  better performance than task-specific models.
We devise SkillNet-X with task-specific skills and language-specific skills and explore two variants to implement SkillNet-X.
We further show that skill pre-training gives a better initialized parameters which further improves the overall performance.
To further investigate the generalization of our model, we adapt our model to two new tasks.
Results show that SkillNet-X could be adapted to new tasks faster (a good starting point) and better (a high evaluation metrics) by combining the learned skills.
We leave the extension to larger number of languages and tasks to the future.

\bibliography{anthology,custom}
\bibliographystyle{acl_natbib}




\end{document}